\documentclass{article}

\usepackage[preprint]{neurips_2024}

\usepackage[utf8]{inputenc} 
\usepackage[T1]{fontenc}    
\usepackage{hyperref}       
\usepackage{url}            
\usepackage{booktabs}       
\usepackage{amsfonts}       
\usepackage{nicefrac}       
\usepackage{microtype}      
\usepackage{xcolor}         
\usepackage{booktabs} 
\usepackage{multirow}
\usepackage{graphicx} 

\definecolor{pastel-red}{HTML}{FF9AA2}
\definecolor{pastel-green}{HTML}{B5EAD7}
\definecolor{pastel-blue}{HTML}{A7DBD8}
\definecolor{pastel-teal}{HTML}{9AD9DB}
\definecolor{pastel-yellow}{HTML}{FFFFCC}
\definecolor{pastel-purple}{HTML}{D3C0F9}
\definecolor{pastel-orange}{HTML}{FFC2A2}
\definecolor{pastel-pink}{HTML}{FFB7B2}
\definecolor{pastel-cyan}{HTML}{A1E8E2}
\definecolor{pastel-lavender}{HTML}{E0BBE4}

\usepackage{fixme}
\fxsetup{status=draft}
\usepackage{color, colortbl}
\usepackage{subcaption}
\usepackage{wrapfig}
\usepackage{amsmath}
\usepackage{cleveref}
\usepackage{color,soul}

\setul{0.5ex}{0.3ex}

\definecolor{Gray}{gray}{0.9}

\title{Mixture-of-Agents Enhances Large Language Model Capabilities}

\author{%
  Junlin Wang \\
  Duke University \\
  Together AI \\
  \texttt{junlin.wang2@duke.edu} \\
  \And
  Jue Wang \\
  Together AI \\
  \texttt{jue@together.ai} \\
  \And
  Ben Athiwaratkun \\
  Together AI \\
  \texttt{ben@together.ai} \\
  \And
  Ce Zhang \\
  University of Chicago \\
  Together AI \\
  \texttt{cez@uchicago.edu} \\
  \And
  James Zou \\
  Stanford University \\
  Together AI\\
  \texttt{jamesz@stanford.edu} \\
}

\begin{document}

\maketitle

\begin{abstract}

Recent advances in large language models (LLMs) demonstrate substantial capabilities in natural language understanding and generation tasks. 
With the growing number of LLMs, how to harness the collective expertise of multiple LLMs is an exciting open direction.
Toward this goal, we propose a new approach that leverages the collective strengths of multiple LLMs through a Mixture-of-Agents (MoA) methodology.
In our approach, we construct a layered MoA architecture wherein each layer comprises multiple LLM agents.
Each agent takes all the outputs from agents in the previous layer as auxiliary information in generating its response.   
MoA models achieves state-of-art performance on  AlpacaEval 2.0, MT-Bench and FLASK, surpassing GPT-4 Omni. For example, our MoA using only open-source LLMs is the leader of AlpacaEval 2.0 by a substantial gap, achieving a score of 65.1\% compared to 57.5\% by GPT-4 Omni.\footnote{
Our code can be found in: \url{https://github.com/togethercomputer/moa}.
}

\end{abstract}

\section{Introduction}
\label{section:introduction}

Large language models (LLMs) \citep{opt,palm,llama,gemini,gpt3,gpt4}
have significantly advanced the field of natural language understanding and generation in recent years.
These models are pretrained on vast amounts of data and subsequently aligned with human preferences to generate helpful and coherent outputs \citep{rlhf}.
However, despite the plethora of LLMs and their impressive achievements, 
they still face inherent constraints on model size and training data.
Further scaling up these models is exceptionally costly, often requiring extensive retraining on several trillion tokens.

At the same time, different LLMs possess unique strengths and specialize in various tasks aspects. 
For instance, some models excel at complex instruction following \citep{wizardlm} 
while others may be better suited for code generation \citep{codellama,deepseekcoder}.
This diversity in skill sets among different LLMs presents an intriguing question: 
\textit{Can we harness the collective expertise of multiple LLMs to create a more capable and robust model?}

\begin{wrapfigure}[15]{r}{0.45\linewidth}
    \centering
    \includegraphics[trim=0 0 0 100,width=1\linewidth]{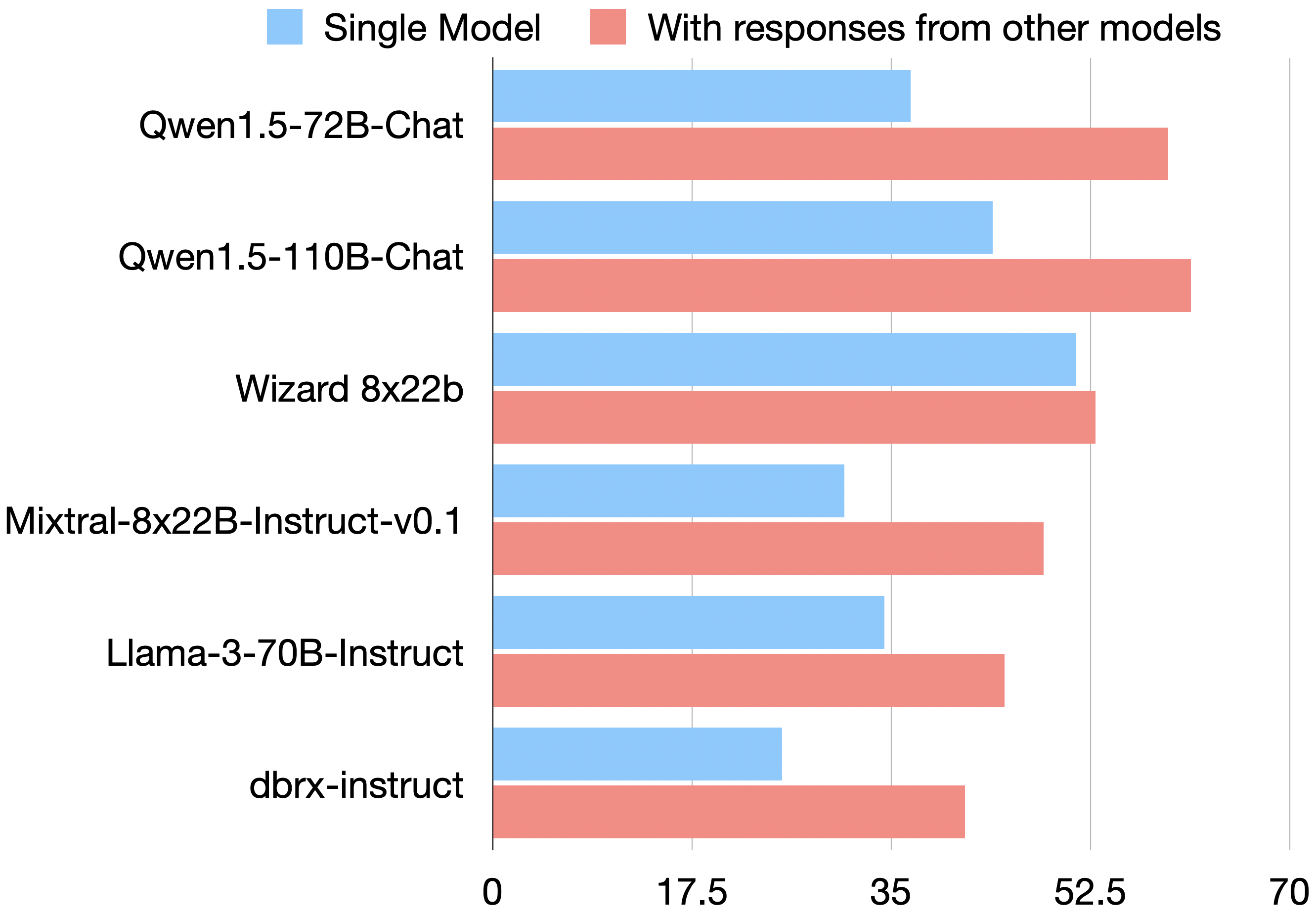}
    \caption{AlpacaEval 2.0 LC win rates improve when provided with responses from other models.}
    \label{fig:collaborativeness}
\end{wrapfigure}

Our answer to this question is \textit{Yes}.
We identify an inherent phenomenon we term the \textit{collaborativeness} of LLMs
— wherein an LLM tends to generate better responses when presented with outputs from other models, even if these other models are less capable by itself. 
\Cref{fig:collaborativeness} showcases the LC win rate on the AlpacaEval 2.0 
benchmark \citep{alpaca-eval} for 6 popular LLMs.
When these models are provided with answers generated independently by these models,
their LC win rates significantly improve.
This indicates that the collaborativeness phenomenon is widespread among LLMs. 
Remarkably, this improvement occurs even when the auxiliary responses provided by the other models are of lower quality than what an individual LLM could generate independently.

\begin{figure*}
    \centering
    \includegraphics[trim=20 0 0 0,width=1.00\linewidth]{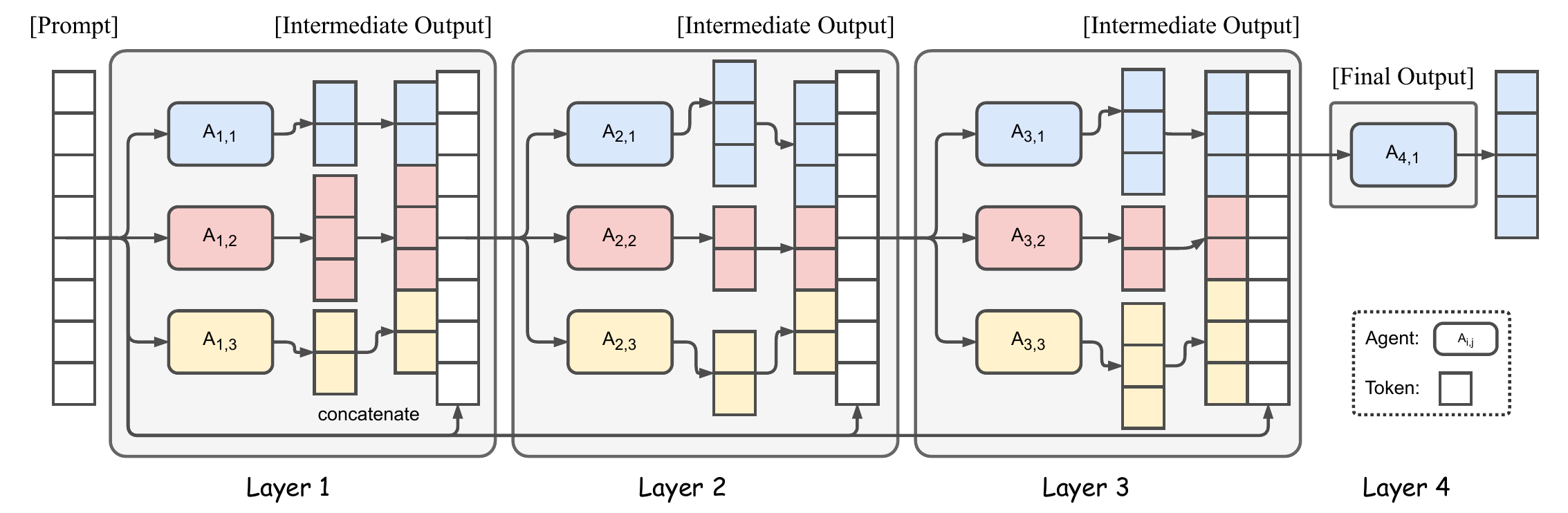}
    \caption{Illustration of the Mixture-of-Agents Structure. 
            This example showcases 4 MoA layers with 3 agents in each layer. 
            The agents here can share the same model.
    }
    \label{fig:mom}
\end{figure*}

Based on this finding, this paper introduces a Mixture-of-Agents (MoA) methodology
that leverages multiple LLMs to iteratively enhance the generation quality.
The structure of MoA is illustrated in \Cref{fig:mom}.
Initially, LLMs in the first layer, denoted as agents $A_{1,1}, ... A_{1,n}$ independently generate responses to a given prompt. 
These responses are then presented to agents in the next layer $A_{2,1}, ... A_{2,n}$ 
(which may reuse a model from the first layer) for further refinement. 
This iterative refinement process continues for several cycles until obtaining a more robust and comprehensive response.

To ensure effective collaboration among models and improve overall response quality, 
careful selection of LLMs for each MoA layer is crucial. 
This selection process is guided by two primary criteria:
(a) Performance Metrics: The average win rate of models in layer \(i\) plays a significant role in determining their suitability for inclusion in layer \(i+1\). Therefore, selecting models based on their demonstrated performance metrics ensures higher-quality outputs.
(b) Diversity Considerations: The diversity of model outputs is also crucial. 
Responses generated by heterogeneous models contribute significantly more than those produced by the same model as we show later in \cref{section:understand_moa}.
By leveraging these criteria — performance and diversity 
— MoA aims to mitigate individual model deficiencies and enhance overall response quality through collaborative synthesis.

We conduct comprehensive evaluations using AlpacaEval 2.0, MT-Bench \citep{mt-bench}, FLASK \citep{flask} benchmarks for assessing the response quality across various dimensions. The results demonstrate substantial improvements with our proposed method, achieving a new SOTA win rate of 65.8\% on AlpacaEval 2.0 compared to the previous best of 57.5\% achieved by GPT-4 Omni.

The \textbf{contributions} of this work are summarized as follows:
(1) \textit{Novel framework}: we propose a Mixture-of-Agents framework designed to leverage the strengths of multiple LLMs, thereby improving their reasoning and language generation capabilities.
(2) \textit{Finding of collaborativeness of language models}: we highlight the inherit collaborativeness among LLMs, where models tend to generate better quality responses when they have access to outputs from other models, even if those outputs are of lower quality.
(3) \textit{State-of-the-art LLM performance}: we conducted extensive experiments using multiple highly-competitive benchmarks such as AlpacaEval 2.0, MT-Bench, and FLASK; our MoA framework achieves state-of-the-art performance on these benchmarks.


\section{Mixture-of-Agents Methodology}
\label{section:method}

In this section, we present our proposed methodology for leveraging multiple models to achieve boosted performance. 
We begin by demonstrating that LLMs possess collaborativeness and thus can improve their responses based on the outputs of other models.
Following this, we introduce the Mixture-of-Agents methodology and discuss its design implications.

\subsection{Collaborativeness of LLMs}

We begin by demonstrating the collaborativeness of LLMs, 
specifically their ability to generate higher quality responses 
when they can reference outputs from other models.
As we have shown in the introduction and \Cref{fig:collaborativeness}, 
many of today's available LLMs exhibit this collaborative capability.

An important pathway to extract maximum benefits from collaboration of multiple LLMs is to characterize how different models are good at in various aspects of collaboration. 
During the collaboration process, we can categorize LLMs into two distinct roles:

\textbf{Proposers} excel at generating useful reference responses for use by other models. 
While a good proposer may not necessarily produce responses with high scores by itself, 
it should offer more context and diverse perspectives, ultimately contributing to better final responses when used by an aggregator.

\textbf{Aggregators} are models proficient in synthesizing responses from other models into a single, high-quality output. 
An effective aggregator should maintain or enhance output quality even when integrating inputs that are of lesser quality than its own.


\Cref{section:role} empirically validate the roles of aggregators and proposers.
Specifically, we show that many LLMs possess capabilities both as aggregators and proposers,
while certain models displayed specialized proficiencies in distinct roles.
GPT-4o, Qwen1.5, LLaMA-3 emerged as a versatile model effective in both assisting and aggregating tasks.
In contrast, WizardLM demonstrated excellent performance as an proposer model 
but struggled to maintain its effectiveness in aggregating responses from other models.

Given that an aggregator can generate higher-quality responses by building upon outputs from other models, 
we propose further enhancing this collaborative potential by introducing additional aggregators. 
One intuitive idea is to replicate the exercise with multiple aggregators 
— initially using several to aggregate better answers and then re-aggregating these aggregated answers.
By incorporating more aggregators into the process, we can iteratively synthesize and refine the responses,
leveraging the strengths of multiple models to produce superior outcomes.
This leads to the design of our proposed Mixture-of-Agents.

\subsection{Mixture-of-Agents}

The structure of MoA is illustrated in \Cref{fig:mom}.
It has \(l\) layers and each layer-\(i\) consists of \(n\) LLMs, denoted by \(A_{i,1}\), \(A_{i,2}\), ..., \(A_{i,n}\).
It is important to note that LLMs can be reused either within the same layer or across different layers. 
When many LLMs in a layer are identical, 
this configuration leads to a special structure that corresponds to a  model generating multiple possibly different outputs (due to the stochasticity of temperature sampling). 
We refer to this setting as single-proposer, where only a sparse subset of models are activated.

Here, each LLM \(A_{i,j}\) processes an input text and generates its continuation.
Our method does not require any fine-tuning and only utilizes the interface of prompting and generation of LLMs. 
Formally, given an input prompt \(x_1\), the output of \(i\)-th MoA layer \(y_i\) can be expressed as follows:
\begin{align}
    y_{i} &= \oplus_{j=1}^n [ A_{i,j}(x_i)] \,\, + x_1, x_{i+1} = y_{i}
    \label{eq:mom}
\end{align}
where $+$ here means concatenation of texts; $\oplus$ means application of the Aggregate-and-Synthesize prompt shown in \Cref{tab:template} to these model outputs. 

\begin{table}[t]
    \centering
    \small
    \caption{Aggregate-and-Synthesize Prompt to integrate responses from other models.}
    \begin{tabular}{@{}p{1\linewidth}@{}}
    \toprule
    You have been provided with a set of responses from various open-source models to the latest user query. Your task is to synthesize these responses into a single, high-quality response. It is crucial to critically evaluate the information provided in these responses, recognizing that some of it may be biased or incorrect. Your response should not simply replicate the given answers but should offer a refined, accurate, and comprehensive reply to the instruction. Ensure your response is well-structured, coherent, and adheres to the highest standards of accuracy and reliability.
    \\ \\
    Responses from models:
    \\
    1. [Model Response from $A_{i,1}$] \\
    2. [Model Response from $A_{i,2}$] \\
    ... \\
    $n$. [Model Response from $A_{i,n}$] \\
    \bottomrule
    \end{tabular}
    \label{tab:template}
\end{table}

In practice, we do not need to concatenate prompt and all model responses so only one LLM is needed to be used in the last layer. 
Therefore, we use the output of an LLM from the $l$-th layer (\( A_{l,1}(x_l) \))
as the final output and evaluate the metrics based on it.

\subsection{Analogy to Mixture-of-Experts}

Mixture-of-Experts (MoE) \citep{moe} is a prominent and well-established technique in machine learning where multiple expert networks specialize in different skill sets. 
The MoE approach has shown significant success across various applications due to its ability to leverage diverse model capabilities for complex problem-solving tasks. 
Our MoA method draws inspiration from this methodology.

A typical MoE design consists of a stack of layers known as MoE layers. 
Each layer comprises a set of \(n\) expert networks alongside a gating network and includes residual connections for improved gradient flow.
Formally, for layer \(i\), this design can be expressed as follows:
\begin{align}
    y_i &= \sum_{j=1}^n {G_{i,j}(x_i) E_{i,j}(x_i)} + x_i 
    \label{eq:moe}
\end{align}
where \(G_{i,j}\) represents the output from the gating network corresponding to expert \(j\), and \(E_{i,j}\) denotes the function computed by expert network \(j\).
The leverage of multiple experts allows the model to learn different skill sets and focus on various aspects of the task at hand.

From a high-level perspective, our proposed MoA framework extends the MoE concept to the model level by operating at the model level rather than at the activation level.
Specifically, our MoA approach leverages LLMs and operates entirely through the prompt interface rather than requiring modifications to internal activations or weights. 
This means that instead of having specialized sub-networks within a single model like in MoE, 
we utilize multiple full-fledged LLMs across different layers.
Note that in our approach, we consolidate the roles of the gating network and expert networks using a LLM, 
as the intrinsic capacity of LLMs allows them to effectively regularize inputs by interpreting prompts and generating coherent outputs without needing external mechanisms for coordination.

Moreover, since this method relies solely on prompting capabilities inherent within off-the-shelf models:
(1) It eliminates computational overhead associated with fine-tuning;
(2) It provides flexibility and scalability: our method can be applied to the latest LLMs regardless of their size or architecture.


\section{Evaluation}
\label{section:evaluation}

This section presents a comprehensive evaluation of our proposed MoA. 
Our findings show that:
\begin{enumerate}
    \item We achieve significant improvements on AlpacaEval 2.0, MT-Bench, and FLASK benchmarks. 
        Notably, with open-source models only, our approach outperforms GPT-4o on AlpacaEval 2.0 and FLASK.
    \item We conduct extensive experiments to provide better understandings of the internal mechanism of MoA.
    \item Through a detailed budget analysis, several implementations of MoA can deliver performance comparable to GPT-4 Turbo while being 2$\times$ more cost-effective.
\end{enumerate}

\subsection{Setup}

\paragraph{Benchmarks} We mainly evaluate models on AlpacaEval 2.0 \citep{alpaca-eval},
a leading benchmark for assessing the alignment of LLMs with human preferences. 
It contains 805 instructions representative of real use cases. 
Each model's response is directly compared against that of the GPT-4 (\texttt{gpt-4-1106-preview}),
with a GPT-4-based evaluator determining the likelihood of preferring the evaluated model's response.
To ensure fairness, the evaluation employs length-controlled (LC) win rates, effectively neutralizing length bias.\footnote{
This metric tracks closely with human preferences, achieving a Spearman correlation of 0.98 with actual human evaluations \citep{alpaca-eval}.}

Additionally, we also evaluate on MT-Bench \citep{mt-bench} and FLASK \citep{flask}.
MT-Bench uses GPT-4 to grade and give a score to model's answer.
FLASK, on the other hand, offers a more granular evaluation with 12 skill-specific scores.

\paragraph{Models} 
In our study, we constructed our default MoA{} by using only open-source models to achieve competitive performance. 
The models included are:
Qwen1.5-110B-Chat \citep{qwen}, Qwen1.5-72B-Chat, 
WizardLM-8x22B \citep{wizardlm}, 
LLaMA-3-70B-Instruct \citep{llama2}, 
Mixtral-8x22B-v0.1 \citep{mixtral}, 
dbrx-instruct \citep{dbrx}.
We construct 3 MoA layers and use the same set of models in each MoA layer.
We use Qwen1.5-110B-Chat as the aggregator in the last layer. 
We also developed a variant called MoA w/ GPT-4o, 
which prioritizes high-quality outputs 
by using GPT-4o as the aggregator in the final MoA layer. 
Another variant, MoA-Lite, emphasizes cost-effectiveness. 
It uses the same set of models as proposers
but includes only 2 MoA layers and employs Qwen1.5-72B-Chat as the aggregator. 
This makes it more cost-effective than GPT-4o 
while achieving a $1.8\%$ improvement in quality on AlpacaEval 2.0.
We ensure strict adherence to the licensing terms of all models utilized in this research.
For open-source models, all inferences were ran  through 
Together Inference Endpoint.\footnote{\url https://api.together.ai/playground/chat}

\subsection{Benchmark Results}


In this subsection, we present our evaluation results on three standard benchmarks:
AlpacaEval 2.0, MT-Bench, and FLASK. 
These benchmarks were chosen to comprehensively assess the performance of our approach and
compare with the state-of-the-art LLMs.

\begin{table}[t]
    \centering
    \small
    \caption{Results on AlpacaEval 2.0 and MT-Bench. For AlpacaEval 2.0, MoA and MoA-Lite correspond to the 6 proposer with 3 layers and with 2 layer respectively. 
    MoA w/ GPT-4o corresponds to using GPT-4o as the final aggregator in MoA.
    We ran our experiments three times and reported the average scores along with the standard deviation.
    $^\dagger$ denotes our replication of the AlpacaEval results.
    We ran all the MT-Bench scores ourselves to get turn-based scores.
    }
    \begin{subtable}{0.42\linewidth}
        \setlength{\tabcolsep}{4pt}
        \centering
        \caption{AlpacaEval 2.0}
        \begin{tabular}{@{}lcc@{}}
        \toprule
        Model               & LC win.   & win. \\
        \midrule
        \rowcolor{Gray}
        MoA w/ GPT-4o            &   65.7$_{\pm \text{0.7}}$\%      & 78.7$_{\pm \text{0.2}}$\% \\
        \rowcolor{Gray}
        MoA{}            &   65.1$_{\pm \text{0.6}}$\%      & 59.8$_{\pm \text{0.3}}$\%          \\
        \rowcolor{Gray}
        MoA-Lite            &   59.3$_{\pm \text{0.2}}$\%      & 57.0$_{\pm \text{0.7}}$\%   \\
        GPT-4 Omni (05/13)  &	57.5\%      & 51.3\%   \\
        GPT-4 Turbo (04/09) &	55.0\%	    & 46.1\%   \\
        WizardLM 8x22B$^\dagger$ & 51.3\%       & 62.3\%       \\
        GPT-4 Preview (11/06) &	50.0\%      & 50.0\% \\
        Qwen1.5 110B Chat   &	43.9\%	    & 33.8\% \\
        Qwen1.5 72B Chat    &	36.6\%	    & 26.5\% \\
        GPT-4 (03/14)       &	35.3\%	    & 22.1\% \\
        Llama 3 70B Instruct &	34.4\%	    & 33.2\% \\
        Mixtral 8x22B v0.1  &	30.9\%	    & 22.2\% \\
        \bottomrule
        \end{tabular}
        \label{tab:alpaca_eval}
    \end{subtable}
    \hfill
    \begin{subtable}{0.56\linewidth}
        \centering
        \setlength{\tabcolsep}{4pt}
        \caption{MT-Bench.}
        \begin{tabular}{@{}lccc@{}}
        \toprule
        Model                   &  Avg.         & 1st turn  & 2nd turn \\
        \midrule
        \rowcolor{Gray}
        MoA w/ GPT-4o           & 9.40$_{\pm \text{0.06}}$       & 9.49          & 9.31   \\
        GPT-4 Turbo (04/09)     & 9.31      	    & 9.35          & 9.28          \\
        \rowcolor{Gray}
        MoA{}                   & 9.25$_{\pm \text{0.10}}$       & 9.44          & 9.07            \\
        GPT-4 Preview (11/06)   & 9.20	            & 9.38          & 9.03         \\
        GPT-4 Omni (05/13)      & 9.19              & 9.31          & 9.07         \\
        \rowcolor{Gray}
        MoA-Lite                & 9.18$_{\pm \text{0.09}}$        & 9.38          & 8.99            \\
        Qwen1.5 110B Chat       & 8.96      	    & 9.23          & 8.63          \\
        Llama 3 70B Instruct    & 8.94      	    & 9.2           & 8.68          \\
        Mixtral 8x22B v0.1      & 8.78      	    & 9.11          & 8.44          \\
        WizardLM 8x22B          & 8.78              & 8.96          & 8.61           \\
        Qwen1.5 72B Chat        & 8.44      	    & 8.55          & 8.34          \\
        GPT-4 (06/13)           & 8.84      	    & 9.08          & 8.61          \\
        \bottomrule
        \end{tabular}
        \label{tab:mt_bench}
    \end{subtable}
\end{table}

\paragraph{AlpacaEval 2.0} 
We conducted comparisons against leading models such as GPT-4 and other state-of-the-art open-source models.
The detailed results are presented in \Cref{tab:alpaca_eval} where our MoA methodology  achieved top positions on the AlpacaEval 2.0 leaderboard, 
demonstrating a remarkable $8.2\%$ absolute improvement over the previous top model, GPT-4o. 
Moreover, it is particularly noteworthy that our model outperformed GPT-4o using solely open-source models, 
achieving a margin of $7.6\%$ absolute improvement from $57.5\%$ (GPT-4o) to $65.1\%$ (MoA).
Our MoA-Lite setup uses less layers and being more cost-effective.
Even with this lighter approach, we still outperform the best model by $1.8\%$, improving from $57.5\%$ (GPT-4o) to $59.3\%$ (MoA-Lite).
This further highlights the effectiveness of our method 
in leveraging open-source models capabilities with varying compute budget to their fullest potential.

\begin{wrapfigure}[22]{R}{0.55\linewidth}
    \centering
     \includegraphics[trim=0 30 0 100, width=\linewidth]{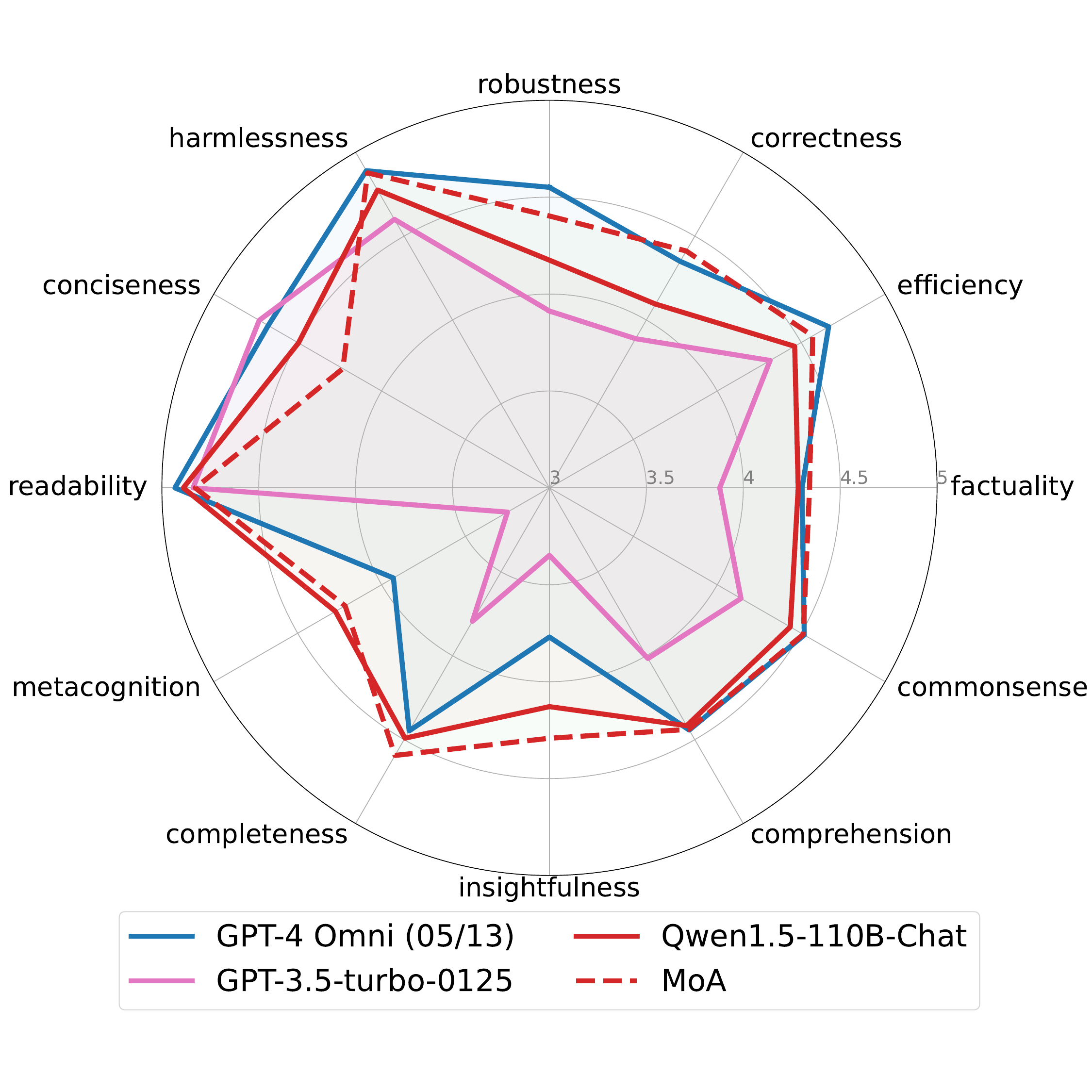}
    \caption{
    Results on FLASK where we use the 6 proposer MoA setup and Qwen1.5-110B-Chat is the aggregator.
    }
    \label{fig:flask}
\end{wrapfigure}

\paragraph{MT-Bench} 
Though improvements over individual models on the MT-Bench are relatively incremental, 
this is understandable given that current models already perform exceptionally well on this benchmark, 
as a single model alone can achieve scores greater than 9 out of 10. 
Despite the marginal enhancements, our approach still secures the top position on the leaderboard. 
This demonstrates that even with already highly optimized benchmarks, 
our method can push the boundaries further, maintaining the leadership.

\paragraph{FLASK} 
FLASK provides fine-grained evaluation of models. 
Among those metrics, MoA excels in several key aspects. 
Specifically, our methodology shows significant improvement in 
robustness, correctness, efficiency, factuality, commonsense, 
insightfulness, completeness, 
compared to the single model score of the aggregator, Qwen-110B-Chat.
Additionally, MoA also outperforms GPT-4 Omni in terms of correctness, factuality, insightfulness, completeness, and metacognition. 
One metric where MoA did not do as well was conciseness; 
the model produced outputs that were marginally more verbose.

\subsection{What Makes Mixture-of-Agents Work Well?}
\label{section:role} \label{section:understand_moa}

In this subsection, we conduct experiments that provide us better understandings of the internal mechanism of Mixture-of-Agents. We summarize  key insights below.

\begin{figure}
    \centering
    \begin{subfigure}[b]{0.495\linewidth}
        \includegraphics[height=5cm]{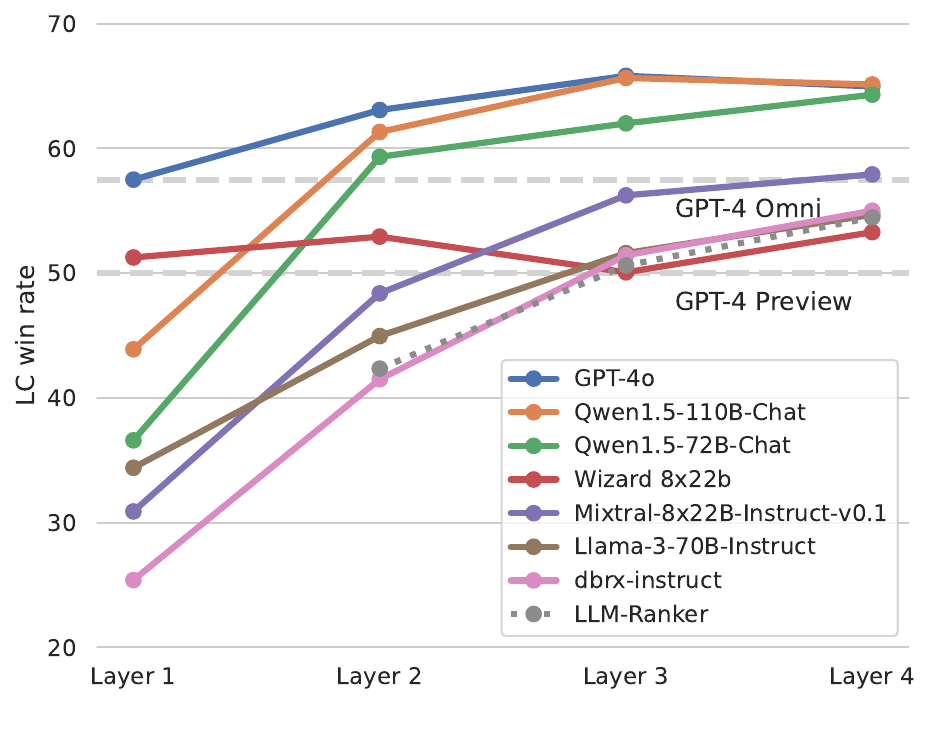}
    \end{subfigure}
    \hfill
    \begin{subfigure}[b]{0.495\linewidth}
        \includegraphics[height=5cm]{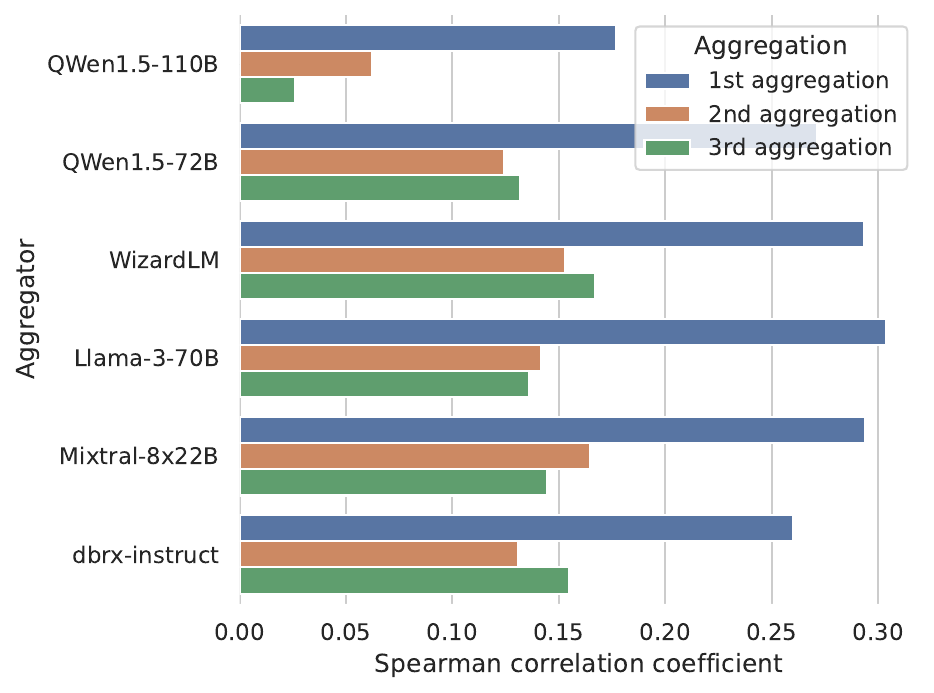}
        \label{fig:h2}
    \end{subfigure}
    \caption{
    (a) LC win rate on AlpacaEval 2.0 with different aggregators in the 6-model Mixture-of-Agents setup. All the curves use the same 6 proposer agents; they only differ in the choice of the final aggregator. The LLM ranker uses Qwen1.5-110B-Chat model with a prompt format in Appendix \Cref{tab:template_ranking}. The GPT-4o model is only used to aggregate the output for the purpose of evaluation and does not participate as a proposer towards the next layer. 
    (b) Spearman correlation between BLEU scores (calculated using 3-gram, 4-gram, and 5-gram metrics) and win rate of the proposed outputs. }
    \label{fig:corr}
    \label{fig:improvements}
\end{figure}

\paragraph{Mixture-of-Agents significantly outperforms LLM rankers.}
 First, we compare Mixture-of-Agents with an LLM-based ranker which uses the aggregator model to select one of the answers that are generated by the proposers, instead of generating a new output. The results are shown in \Cref{fig:improvements}, where we can observe that the MoA approach significantly outperforms an LLM-ranker baseline. The fact that MoA outperforms the ranking approach suggests that the aggregator does not simply select one of the generated answers by the proposers, but potentially performs sophisticated aggregation over all proposed generations.

\paragraph{MoA tends to incorporate the best proposed answers.}
 We also compare the aggregator's response with the proposers' responses via similarity scores such as BLEU \citep{bleu} which reflects n-gram overlaps. Within each sample, given $n$ proposed answers by the proposers, we calculate the the Spearman's rank correlation coefficient between $n$ similar scores and $n$ preference scores determined by the GPT-4 based evaluator.
 The results in \Cref{fig:corr} indeed confirms a positive correlation between the win rate and the BLEU score. 
 We also provide results with Levenshtein similarity \citep{RapidFuzzPythonLevenshtein} or TF-IDF as opposed to BLEU scores in \Cref{section:corr_more}.
 where both alternative approaches for textual similarities also yield positive correlation with the preference scores.

\paragraph{Effect of model diversity and the number of proposers.}

We analyze how the number of proposals affect the final output quality by varying $n$, the number of proposers in each layer. 
We show the results in \Cref{tab:ablation} where we find that scores increases monotonically with $n$, reflecting the benefits of having more auxiliary information. In addition, we also quantify  the impact of using a diverse set of LLMs as proposers. For each $n$, we compare two settings: ``single-proposer'' where the $n$ responses are generated by the same LLM with 
a temperature of 0.7;
and ``multiple-proposer'' where each response is generated by a different LLMs. Overall, using multiple different LLMs consistently yielded better results. Both results suggest that having a larger number of diverse LLM agents in each MoA layer can improve performance. Further scaling the width of MoA is a promising direction of future investigation.




\paragraph{Specialization of models in the Mixture-of-Agent ecosystem.}
We also conducted experiments to determine which models excel in specific roles.
Specifically, \Cref{tab:proposers_vs_aggregators} shows that GPT-4o, Qwen, LLaMA-3 emerged as a versatile model effective in both assisting and aggregating tasks.
In contrast, WizardLM demonstrated excellent performance as an proposer model 
but struggled to maintain its effectiveness in aggregating responses from other models.

\begin{table}[t]
    \begin{minipage}[t]{0.49\linewidth}
        \centering
        \small
        \caption{Effects of the number of proposer models on AlpacaEval 2.0. 
                  We denote $n$ as either the number of agents in an MoA layer or the number of proposed outputs in the single-proposer setting. 
                  We use Qwen1.5-110B-Chat as the aggregator and use 2 MoA layers for all settings in this table.
        }
        \begin{tabular}{@{}lcc@{}}
        \\
        \toprule  
        Setting               & Multiple-Proposer  &     Single-Proposer  \\
        \midrule 
        $n=6$               & 61.3\%   & 56.7\%           \\
        $n=3$               & 58.0\%   & 56.1\%           \\
        $n=2$               & 58.8\%   & 54.5\%          \\
        $n=1$               & 47.8\%   & 47.8\%          \\
        \bottomrule 
        \end{tabular}
        \label{tab:ablation}
    \end{minipage}
    \hfill
    \begin{minipage}[t]{0.49\linewidth}
        \setlength{\tabcolsep}{4pt}
        \centering
        \small
        \caption{
        Impact of different models serving as proposers vs aggregators. 
        When evaluating different aggregators, all six models serve as proposers;
        when evaluating proposers, Qwen1.5-110B-Chat serves as the aggregator. 
        We use 2 MoA layers in this table.
        }
        \begin{tabular}{@{}lcc@{}}
        \\
            \toprule
            Model                   & As aggregator & As proposer  \\
            \midrule
            Qwen1.5-110B-Chat       &   61.3\%     & 56.7\%          \\
            Qwen1.5-72B-Chat        &   59.3\%     & 53.3\%          \\
            LLaMA-3-70b-Instruct    &   45.0\%     & 60.6\%           \\
            WizardLM 8x22B          &   52.9\%     & 63.8\%          \\
            Mixtral-8x22B-Instruct  &   48.4\%     & 54.8\%          \\
            dbrx-instruct           &   41.5\%     & 55.1\%         \\
            \bottomrule
            \end{tabular}
        \label{tab:proposers_vs_aggregators}
    \end{minipage}
\end{table}

\subsection{Budget and Token Analysis}
\label{section:budget}

\begin{figure}
    \centering
    \begin{subfigure}[b]{0.49\linewidth}
        \includegraphics[width=\linewidth]{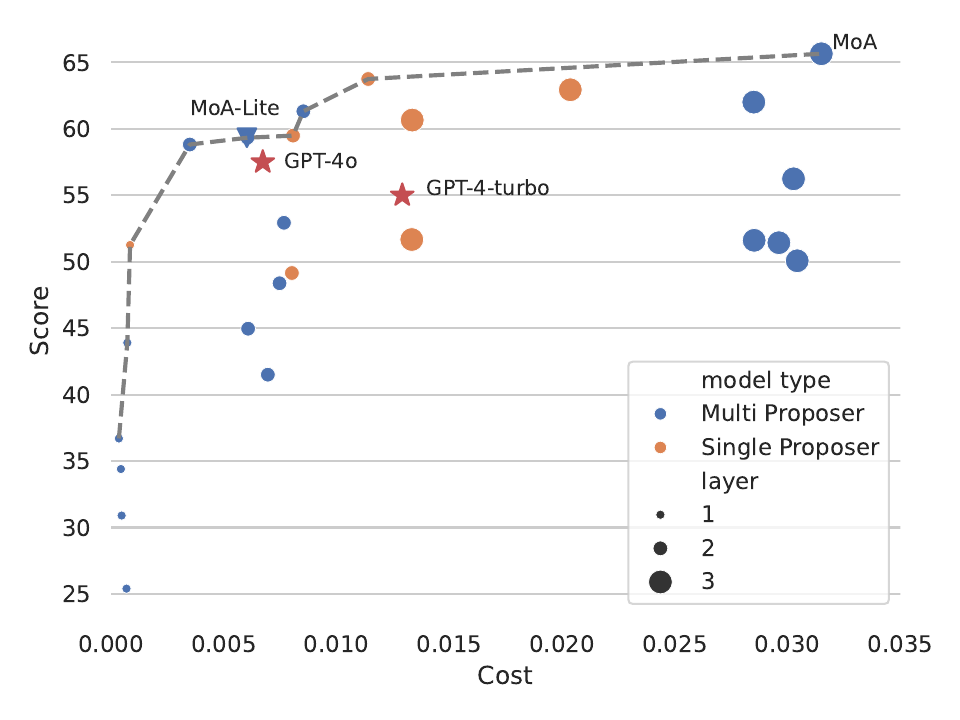}
        \caption{LC win rate vs. cost}
        \label{fig:cost}
    \end{subfigure}
    \hfill
    \begin{subfigure}[b]{0.49\linewidth}
        \includegraphics[width=\linewidth]{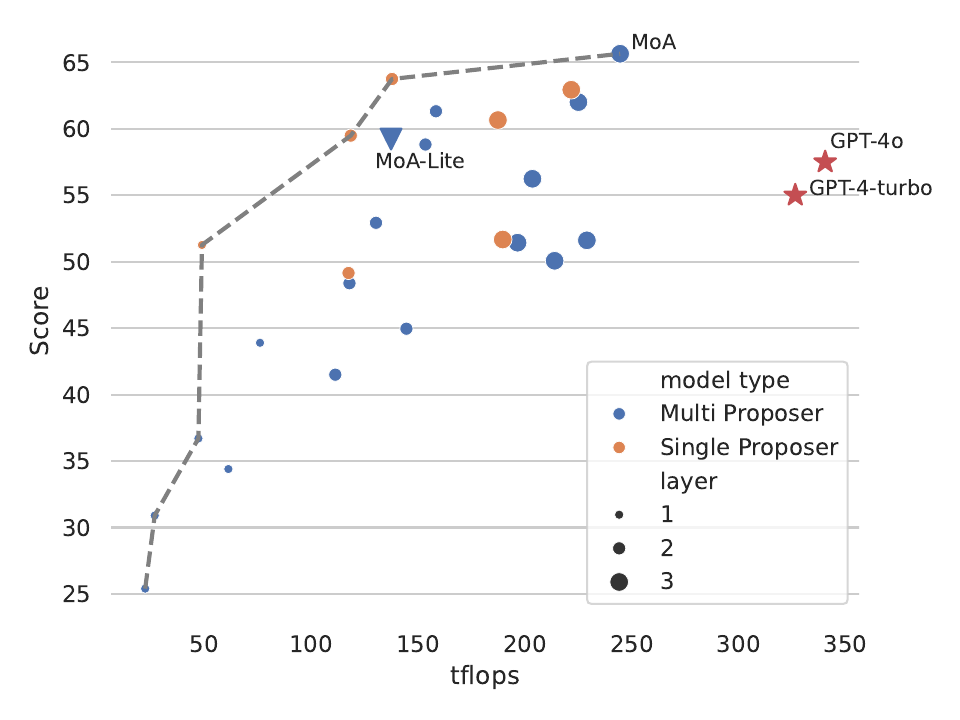}
        \caption{LC win rate vs. tflops}
        \label{fig:tokens}
    \end{subfigure}
    \caption{(a) Performance trade-off versus cost. (b) Performance trade-off versus the number of tera floating operations (tflops), which we use as a proxy for latency. Note that we calculate the sum over layers of the max number of tflops among proposers in each MoA layer as multiple proposers can run in parallel.
    Our plots illustrate a Pareto frontier where we can choose a model progressively higher score with the lowest cost for such level of performance. 
    We show that the Mixture-of-Agents approach lie on this Pareto front, 
    as opposed to GPT-4 Turbo and GPT-4o which are not cost optimal and is more expensive compared to MoA approaches of the same LC win rate. 
    \textit{Single Proposer}: uses the same model to generate multiple responses in each MoA layer;
    \textit{Multi Proposer}: uses different models in each MoA layer. The actual tflops of GPT-4 is unknown, so we use the rumored size from the community of an 8x220B architecture.
    }
\end{figure}


To understand the relationship between budget, token usage, and LC win rates, 
we conducted a budget and token analysis. 
\Cref{fig:cost} and \Cref{fig:tokens} illustrate these relationships.

\paragraph{Cost Effectiveness}
In \Cref{fig:cost}, we plot the LC win rate against 
the average inference cost for each instance in the AplacaEval 2.0 benchmark.
The cost is calculated based on pricing information available from API provider websites.\footnote{
For open-source models, we calculate the price using data from \url{https://api.together.ai/models};
for OpenAI models, we use pricing details from \url{https://openai.com/api/pricing/}. 
Pricing data was retrieved as of May 22, 2024.
}
This helps identify cost-effective models that achieve high performance without incurring excessive expenses.
The chart reveals a Pareto front where certain models strike an optimal balance between cost and performance.
Models closer to this Pareto front are more desirable as they provide better monetary value by delivering high LC win rates at lower costs.
Specifically, if we prioritize the quality, MoA is the best configuration.
However, if we want to strike a good balance between quality and cost, MoA-Lite can match GPT-4o's cost while achieving higher level of quality.
Notably, it outperforms GPT-4 Turbo by approximately $4\%$ while being more than twice as cost-effective.

\paragraph{Tflops Consumption}
\Cref{fig:tokens} depicts the relationship between LC win rate and the number of tflops. 
Here we use the number of tflops as a proxy for latency since latency can vary depending on the inference systems.
This analysis is crucial for understanding how different models manage their budgets
while maintaining or improving performance levels. 
Similar to the cost efficiency analysis, a Pareto front can be observed here as well.
Models on this front effectively utilize their computational resource to maximize their LC win rate.


\section{Related Work}
\label{section:related_work}

\subsection{LLM Reasoning}

In order to improve generation quality of LLMs, 
recent researches have experienced great progresses in optimizing LLMs to various downstream tasks through prompt engineering.
Chain of Thought (CoT) \citep{cot, cot0} prompting techniques represent a linear problem-solving approach where each step builds upon the previous one.
\citet{firstms} applied CoT to multi-step reasoning tasks. 
To automate CoT prompting, Auto-CoT \citep{autop} constructs demonstrations by sampling diverse questions and generating reasoning chains. 
Active-Prompt \citep{activep}  focuses on selecting the most uncertain questions for task-specific annotations.
PS Prompt \citep{pas} decomposes tasks into subtasks.
Tree-of-Thought (ToT) \citep{tot} expands on the reasoning process by considering multiple paths of reasoning and self-evaluating choices.
Effective Graph-of-Thought \citep{eGoT} frames thoughts as graphs.
Natural Program prompting \citep{np} is proposed for better solving deductive reasoning tasks.
And re-reading prompt \citep{reread} revisits question information embedded within input prompts.

\subsection{Model Ensemble}

A straightforward solution to leverage the strengths of multiple models is reranking outputs from different models.
For instance, \citet{jiang-etal-2023-llm} introduce \textsc{PairRanker}, 
which performs pairwise comparisons on candidate outputs to select the best one, showing improvements on a self-constructed instruction dataset. 
To address the substantial computational costs associated with multi-LLM inference, 
other studies have explored training a \textit{router} that predicts the best-performing model from a fixed set of LLMs for a given input~\citep{wang2024fusing,shnitzer2024large,lu2023routing}.
Additionally, FrugalGPT \citep{chen2023frugalgpt} proposed reducing the cost of using LLMs by employing different models in a cascading manner.
In order to better leverage the responses of multiple models,
\citet{jiang-etal-2023-llm} trained a \textsc{GenFuser}, 
a model that was trained to generate an improved response to capitalize on the strengths of multiple candidates.
\citet{huang2024enabling} proposed to fuse the outputs of different models by averaging their output probability distributions.

Another line of work is multi-agent collaboration.
Several studies explore using multiple large language models as agents that collectively discuss and reason through given problems interactively.
\citet{debate} establishes a mechanism for symmetric discussions among agents. 
Around the same time, MAD \citep{mad} introduces an asymmetric mechanism design, with different roles, i.e., debater and judge. Other similar works include \citep{chan2023chateval}. 
Moreover, ReConcile \citep{reconcile} exemplifies an asymmetric discussion involving weighted voting. 
To understand discussion more deeply, \citet{expd} aim to explain such collaboration mechanism in a social psychology view. 
\citet{wang2024rethinking} systematically compared multi-agent approaches and found a single agent with a strong prompt including detailed demonstrations can achieve comparable response quality to multi-agent approaches. 


\section{Conclusion}
\label{section:limitation}
\label{section:broad}

This paper introduces a Mixture-of-Agents approach aimed at leveraging the capabilities of multiple LLMs via successive stages for iterative collaboration. Our method harnesses the collective strengths of agents in the Mixture-of-Agents family, and can significantly improve upon the output quality of each individual model.
Empirical evaluations conducted on AlpacaEval 2.0, MT-Bench, and FLASK demonstrated substantial improvements in response quality, with our approach achieving the LC win rate up to $65\%$. 
These findings validate our hypothesis that integrating diverse perspectives from various models can lead to superior performance compared to relying on a single model alone. In addition, we provide insights into improving the design of MoA; systematic optimization of MoA architecture is an interesting direction for future work. 


\paragraph{Limitations.}
Our proposed method requires iterative aggregation of model responses, 
which means the model cannot decide the first token until the last MoA layer is reached. 
This potentially results in a high Time to First Token (TTFT), which can negatively impact user experience. 
To mitigate this issue, we can limit the number of MoA layers, 
as the first response aggregation has the most significant boost on generation quality.
Future work could explore chunk-wise aggregation instead of aggregating entire responses at once, which can reduce TTFT while maintaining response quality.

\paragraph{Broader Impact.}
This study holds the potential to enhance the effectiveness of LLM-driven chat assistants, thereby making AI more accessible. 
Moreover, since the intermediate outputs that are expressed in natural language, MoA presented improves the interpretability of models.
This enhanced interpretability facilitates better alignment with human reasoning.


\newpage
\bibliographystyle{neurips24}
\bibliography{neurips24}


\appendix

\section*{Supplementary Material}

\section{Spearman Correlation using Different Similarity Functions}
\label{section:corr_more}

\begin{figure}
    \centering
    \begin{subfigure}[b]{0.495\linewidth}
        \includegraphics[height=5cm]{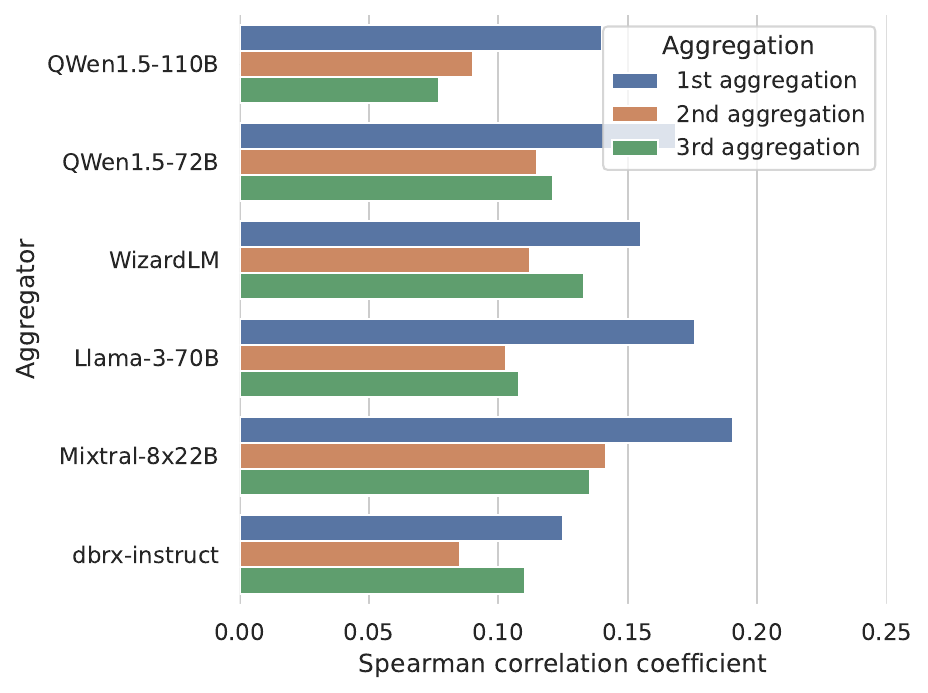}
    \end{subfigure}
    \hfill
    \begin{subfigure}[b]{0.495\linewidth}
        \includegraphics[height=5cm]{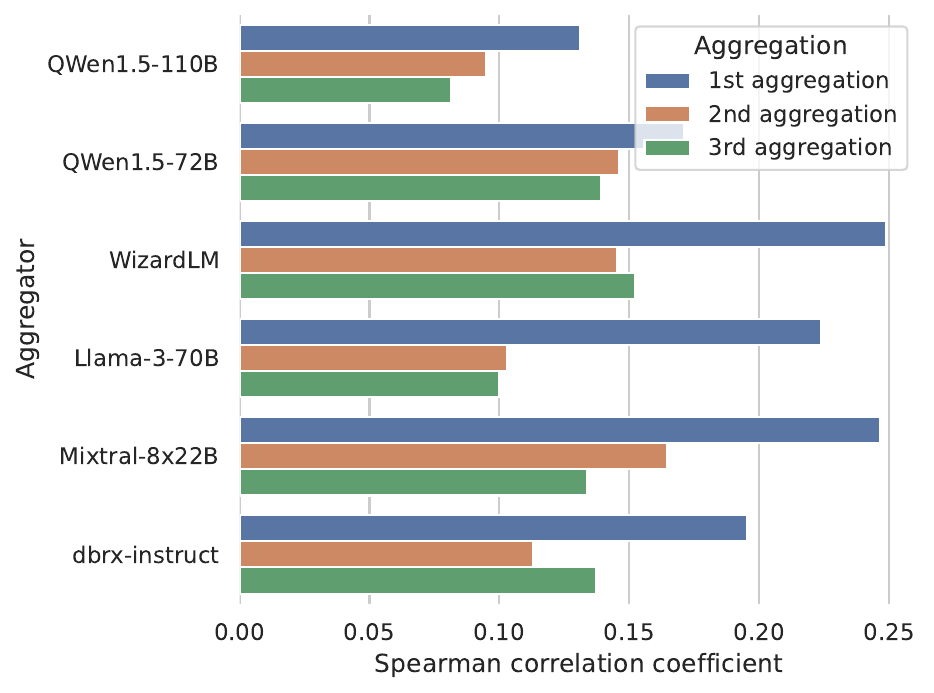}
        \label{fig:h2}
    \end{subfigure}
    \caption{(a) Spearman Correlation using TF-IDF similarity; (b) Spearman Correlation using Levenshtein similarity.}
    \label{fig:corr_more}
\end{figure}

We present results using TF-IDF-based similarity and Levenshtein similarity when calculating the Spearman correlation. 
Specifically, within each sample of \( n \) proposed answers, 
we calculate Spearman correlation coefficient between the \( n \) similarity scores and the \( n \) preference scores determined by the GPT-4-based evaluator. 
As shown in \Cref{fig:corr_more}, there is indeed a positive correlation between win rate and both TF-IDF similarity and Levenshtein similarity.

\section{LLM Ranker}

This section introduces the setup of the LLM-Ranker used in this paper. 
The LLM-Ranker is designed to evaluate and rank the best output generated by some LLMs.
\Cref{tab:template_ranking} presents the template for prompting the model during these evaluations. 
We use this LLM-Ranker to pick the best answer among and use AlpacaEval evaluator 
to evaluate the best ranked answer.

\begin{table}[t]
    \centering
    \small
    \caption{Prompt for ranking with LLMs}
    \begin{tabular}{@{}p{1\linewidth}@{}}
    \toprule
    
You are a highly efficient assistant, who evaluates and selects the best large language model (LLMs) based on the quality of their responses to a given instruction. This process will be used to create a leaderboard reflecting the most accurate and human-preferred answers.
\\
I require a leaderboard for various large language models. I'll provide you with prompts given to these models and their corresponding outputs. Your task is to assess these responses, and select the model that produces the best output from a human perspective.
\\
\\
\#\# Instruction
\\
\begin{verbatim}
{
    "instruction": """{instruction}""",
}
\end{verbatim}
\\
\#\# Model Outputs
\\
Here are the unordered outputs from the models. Each output is associated with a specific model, identified by a unique model identifier.

\begin{verbatim}
{
    {
        "model_identifier": "{identifier_1}",
        "output": """{output_1}"""
    },
    {
        "model_identifier": "{identifier_2}",
        "output": """{output_2}"""
    },
    {
        "model_identifier": "{identifier_3}",
        "output": """{output_3}"""
    },
    {
        "model_identifier": "{identifier_4}",
        "output": """{output_4}"""
    },
    {
        "model_identifier": "{identifier_5}",
        "output": """{output_5}"""
    },
    {
        "model_identifier": "{identifier_6}",
        "output": """{output_6}"""
    }
}
\end{verbatim}
\\
\\
\#\# Task
\\
Evaluate the models based on the quality and relevance of their outputs, and select the model that generated the best output. Answer by providing the model identifier of the best model. We will use your output as the name of the best model, so make sure your output only contains one of the following model identifiers and nothing else (no quotes, no spaces, no new lines, ...).
\\
\\
\#\# Best Model Identifier
\\
    \bottomrule
    \end{tabular}
    \label{tab:template_ranking}
\end{table}

\section{Case Study}

\begin{table}[t]
    \centering
    \small
    \caption{Case: Some models produce high quality answers.}
    \begin{tabular}{@{}llp{0.6\linewidth}@{}}
    \toprule
    Role        &  Preference   & Content \\
    \midrule
    User        &      -         & I want to talk about "Smooth",  a song by Rob Thomas. \\
    \midrule
    Qwen1.5-110B-Chat            & 0.35
    &  ... \setulcolor{blue}\ul{It's a perfect blend of rock, pop, and Latin music, showcasing Santana's signature guitar work and Rob Thomas's soulful vocals} ...
       
    \\   
    Qwen1.5-72B-Chat             & 0.00
    &  ...
        
    \\     
    Llama-3-70B-Instruct         & 0.00
    & ... \setulcolor{red}\ul{it's a collaboration between Rob Thomas, the lead vocalist of Matchbox Twenty, and Santana, the legendary guitarist} ...
    
    \\         
    WizardLM-2-8x22B             & 0.99
    & ... \setulcolor{orange}\ul{It spent 12 weeks at the top of the Billboard Hot 100 chart in the United States} ...
         \setulcolor{orange}\ul{capturing the song's sultry and energetic vibe} ...
    
    \\     
    Mixtral-8x22B-Instruct-v0.1  & 0.00
    & ...
    
    \\        
    dbrx-instruc                 & 0.00
    & ...
    
    \\ 
    \midrule
    \begin{tabular}{@{}l@{}}
    Aggregated \\
    (Qwen1.5-110B-Chat)
    \end{tabular}
    & 0.99
    & "Smooth" is actually \setulcolor{red}\ul{a collaboration between Rob Thomas, who is the lead singer of the band Matchbox Twenty, and the renowned Latin rock guitarist Carlos Santana} ... \setulcolor{blue}\ul{This iconic song blends rock, pop, and Latin music elements, showcasing Santana's signature guitar work and Rob Thomas's distinctive vocals} ... \setulcolor{orange}\ul{The song was a colossal success, spending 12 weeks at the top of the Billboard Hot 100} ... The music video, \setulcolor{orange}\ul{set in a vibrant club scene, complements the song's sultry vibe} ...
    
    \\ 
    \bottomrule
    \end{tabular}
    \label{tab:case1}
\end{table}

\begin{table}[t]
    \centering
    \small
    \caption{Case: all proposed responses are not good enough.}
    \begin{tabular}{@{}llp{0.6\linewidth}@{}}
    \toprule
    Role        &  Preference   & Content \\
    \midrule
    User        &      -         & How do you become an author? \\
    \midrule
    Qwen1.5-110B-Chat            & 0.00 
    &  Becoming an author involves several steps and ...
    3. \setulcolor{blue}\ul{**Choose your niche**}:...
    8. \setulcolor{blue}\ul{**Understand the publishing industry**}:...
    10. \setulcolor{blue}\ul{**Self-publishing options**}:...
       
    \\   
    Qwen1.5-72B-Chat             & 0.00
    &  ...
        
    \\     
    Llama-3-70B-Instruct         & 0.16
    & Becoming an author can be a rewarding and fulfilling career...
    1. \setulcolor{red}\ul{**Develop a love for reading and writing**}:...
    10. \setulcolor{red}\ul{**Market and promote your work**}:...
    11. \setulcolor{red}\ul{**Continuously improve**}:...
    
    \\         
    WizardLM-2-8x22B             & 0.03
    & 1. ...
    4. \setulcolor{orange}\ul{Finish your work}:...
    
    \\     
    Mixtral-8x22B-Instruct-v0.1  & 0.00
    & ...
    
    \\        
    dbrx-instruc                 & 0.00
    & ...
    
    \\ 
    \midrule
    \begin{tabular}{@{}l@{}}
    Aggregated \\
    (Qwen1.5-110B-Chat)
    \end{tabular}
    & 0.33
    & Becoming an author involves a dedicated journey...
    1. \setulcolor{red}\ul{**Cultivate a Love for Writing and Reading**}:...
    4. \setulcolor{blue}\ul{**Choose Your Niche**}:...
    5. \setulcolor{orange}\ul{**Finish Your Work**}:...
    10. \setulcolor{blue}\ul{**Self-Publishing**}:...
    11. \setulcolor{red}\ul{**Marketing and Promotion**}:...
    12. \setulcolor{red}\ul{**Continuous Learning and Writing**}:...
    
    \\ 
    \bottomrule
    \end{tabular}
    \label{tab:case2}
\end{table}

We present a case study in this section. 
Due to the length of the responses generated by all models, we will only show selected fragments for brevity. 
To illustrate how the aggregator synthesizes the response, 
we underlined similar expressions between the proposed responses and the aggregated response in different colors.
We omit the content that all proposed responses have mentioned.

\Cref{tab:case1} showcases the responses generated by different proposers.
The aggregated response generated by Qwen1.5-110B-Chat 
reflects a high preference for its own content but also incorporates key points 
from Llama-3-70B-Instruct and WizardLM 8x22B. 
Notably, GPT-4's preference score for WizardLM 8x22B's response is 0.99, 
and the final aggregated answer also achieves a preference score of 0.99.

Meanwhile, \Cref{tab:case2} presents another case
where none of the proposed responses achieve a high GPT-4 preference score. 
Despite this, the aggregator successfully identifies and 
incorporates the strong points from these responses,
achieving a preference score of 0.33.

\section{MATH Task}

Here, we demonstrate that our approach is applicable to reasoning tasks, 
such as those in the MATH dataset \cite{hendrycks2021measuring}.
The results are presented in \Cref{tab:MATH}, where we show that our method consistently enhances accuracy by a significant margin. 
This indicates that our approach is also effective for this type of task.
Notably, our method is complementary to existing reasoning techniques such as Chain of Thought \cite{cot} and Self-consistency \cite{cot-sc}.

\begin{table}[t]
\centering
\caption{
Results on the MATH task. 
We evaluate different aggregators,
with all six models serving as proposers in each MoA layer.
}
\begin{tabular}{@{}lllll@{}}
\toprule
Aggregator                  & Layer 1 & Layer 2 & Layer 3  \\
\midrule
Qwen1.5-72B-Chat            & 0.428              & 0.526   & 0.552    \\
Qwen1.5-110B-Chat           & 0.500              & 0.570   & 0.576    \\
Wizard 8x22b                & 0.544              & 0.574   & 0.580    \\
Mixtral-8x22B-Instruct-v0.1 & 0.282              & 0.534   & 0.556    \\
Llama-3-70B-Instruct        & 0.456              & 0.584   & 0.578    \\
dbrx-instruct               & 0.314              & 0.456   & 0.522    \\
\bottomrule
\end{tabular}
\label{tab:MATH}
\end{table}

\end{document}